\documentclass{iopjournal}

\usepackage{xcolor}
\usepackage[normalem]{ulem}
\usepackage{natbib}
\usepackage{lmodern}
\usepackage{hyperref}
\usepackage{url}
\usepackage{booktabs}
\usepackage{amsfonts}       
\usepackage{nicefrac}       
\usepackage{microtype}      
\usepackage{makecell}
\usepackage{graphicx}
\usepackage{wrapfig}
\usepackage{amsmath}
\usepackage{bm}
\usepackage{multirow}
\usepackage[table]{xcolor}
\usepackage[title]{appendix}
\usepackage{float}
\newcommand{\std}[1]{\small{$\pm$#1}}
\begin{document}

\articletype{Paper} 

\title{Cortical-SSM: A Deep State Space Model for \\ Motor Imagery Decoding from EEG Signals}

\author{Shuntaro Suzuki*\orcid{0009-0008-5564-3835}, Shunya Nagashima\orcid{0009-0007-9741-7628} and Komei Sugiura\orcid{0000-0002-0261-0510}}

\affil{Keio University\\
*Author to whom any correspondence should be addressed.
}

\email{shuntaro20021227@keio.jp}

\keywords{Brain-Computer Interface, Electroencephalography, Motor Imagery, Deep State Space Model}

\begin{abstract}
\textit{Objective.} Classification of  electroencephalogram (EEG) signals obtained during motor imagery (MI) has substantial application potential, including for communication assistance and rehabilitation support for patients with motor impairments. 
These signals remain inherently susceptible to physiological artifacts (e.g., eye blinking, swallowing), which pose persistent challenges.
Although Transformer-based approaches for classifying EEG signals have been widely adopted, they often struggle to capture fine-grained dependencies within them.
\textit{Approach.} To overcome these limitations, we propose Cortical-SSM, a novel architecture that extends deep state space models to capture integrated dependencies of EEG signals across temporal, spatial, and frequency domains. 
We validated our method across two large-scale public MI EEG datasets containing more than 50 subjects.
\textit{Main results.}
Our method outperformed baseline methods on the two benchmarks.
Furthermore, visual explanations derived from our model indicate that it effectively captures neurophysiologically relevant regions of EEG signals.
\textit{Significance.}
These results indicate that Cortical-SSM provides a robust and interpretable alternative to attention-based architectures for MI EEG decoding. 
By enabling physiologically grounded feature learning, our method advances the reliability of subject-independent EEG classification and supports the development of practical and clinically deployable brain–computer interface systems.

\end{abstract}

\section{Introduction}

Brain-Computer Interfaces (BCIs) hold transformative potential across various domains, including the diagnosis of neurodegenerative diseases~\cite{wolpaw2013}, advanced brain function mapping, robotic control, and the development of immersive gaming devices~\cite{hramov2021}.
Among the various BCI paradigms, motor imagery (MI) BCIs decode intentionally modulated neural activity resulting from conscious cognitive effort, making them especially promising for developing assistive communication systems and neurorehabilitation protocols for patients with severe motor impairments~\cite{hramov2021}.
Furthermore, BCI implementations leverage various techniques for recording brain activity, including electroencephalography (EEG), electrocorticography, functional magnetic resonance imaging, and functional near-infrared spectroscopy~\cite{ramadan2017}.
Among these modalities, BCIs based on EEG are particularly promising for real-world applications because of their high temporal resolution and superior portability.

\begin{figure}[h]
    \centering
    \includegraphics[width=\linewidth]{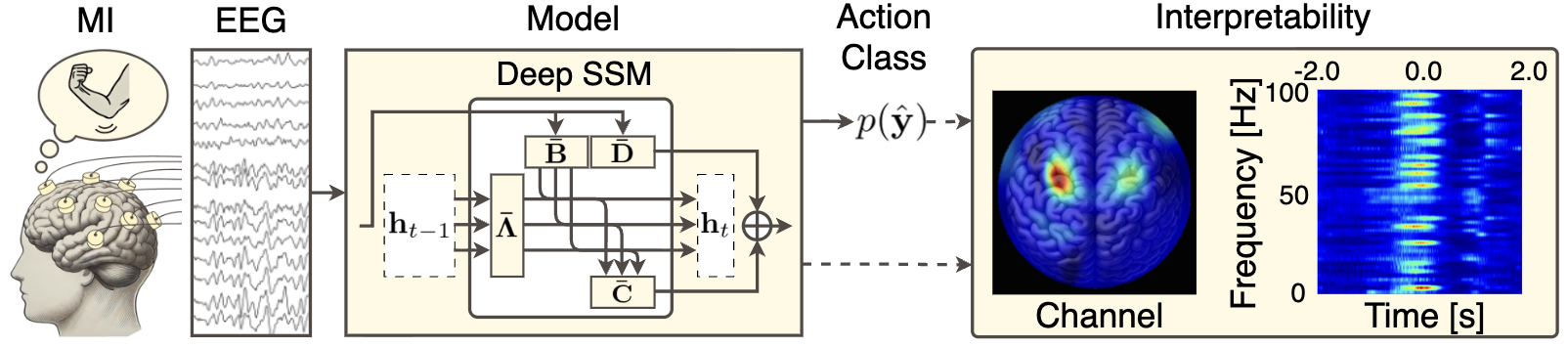}
    \vspace{-2em}
    \caption{
    Task overview. The input is EEG signals recorded while the subject imagines actions (e.g., elbow extension), and the model's output is predictions of the corresponding actions. 
    The model also provides interpretable spatio-temporal and temporal-frequency visualizations.
    }
    \label{fig:eye_catch}
    \vspace{-0.5em}
\end{figure}

In this study, we focus on brain activity occurring during motor imagery (MI) and tackle a classification task for imagined actions using EEG signals.
Figure~\ref{fig:eye_catch} presents a representative example of the task. 
In this case, the input comprises EEG signals recorded while the subject imagines an elbow extension. 
As its response, the model outputs predicted probabilities for each corresponding action.
Despite intensive research, the accurate decoding of MI EEG signals is still non-trivial.
For instance, a binary classification task using MI EEG signals from the OpenBMI~\cite{OpenBMI} dataset, a representative model (e.g., \cite{EEGNet}), demonstrates an error rate of approximately 22\%.

Recent approaches have used Transformers to capture subject-agnostic dependencies in EEG signals (e.g., EEG Conformer~\cite{EEGConformer} and Medformer~\cite{Medformer}).
However, to address the exponential growth in computational cost associated with longer time-series, these methods patchify and compress the input EEG signals along the temporal dimension, which may result in fine-grained temporal dependencies being lost. 
Moreover, while these approaches offer interpretability in spatio-temporal domains, they do not directly provide insights into the frequency domain, leaving the contributions of neurophysiologically significant frequency bands (e.g., mu band~\cite{pfurtscheller2006}) unclear.

Therefore, we propose Cortical-SSM, an extension of Deep SSM that explicitly captures integrated EEG dependencies across temporal, spatial, and frequency domains.
This design enables the modeling of temporal dependencies in EEG signals without compressing them. 
Moreover, the proposed method provides direct visual explanations across the temporal, spatial, and frequency domains, in contrast to prior methods that operate on a subset of domains.

In Cortical-SSM, we extend a deep state space model (Deep SSM~\cite{S5})—an architecture offering superior computational efficiency to Transformers for long sequence modeling—to effectively capture multi-scale temporal dependencies in EEG signals. 
Specifically, we introduce the Frequency-SSM module, which extracts spatio-temporal dependencies for each frequency component, and the Channel-SSM module, which captures temporal-frequency dependencies for each electrode. 
Furthermore, we introduce a Wavelet-Convolution module integrating both deterministic and adaptive frequency features for feature extraction from the frequency domain.
The Frequency-SSM and Channel-SSM modules model temporal dependencies in EEG signals without requiring the patchification employed in prior studies, thereby facilitating the capture of fine-grained temporal variations. 
Moreover, by incorporating the Wavelet-Convolution, Cortical-SSM extracts frequency-analyzable features while simultaneously learning their representations. 
Frequency-SSM and Channel-SSM modules explicitly model frequency-wise and electrode-wise features derived from the Wavelet-Convolution module, enabling direct visual explanations across temporal, spatial, and frequency domains.

Our main contributions lie in the following aspects:
\begin{itemize}
    \item[$\bullet$] We propose Cortical-SSM, an extension of Deep SSM that captures integrated EEG dependencies across temporal, spatial, and frequency domains.
    \item[$\bullet$] For frequency domain feature extraction, we introduce Wavelet-Convolution, which integrates deterministically obtained frequency components with adaptively derived frequency features. This approach enables the extraction of interpretable features while preserving their learnable representations. 
\end{itemize}

\section{Related Work}
Deep learning-based methods for EEG decoding have been extensively studied, as reviewed by Abibullaev et al.~\citep{abibullaev2023} and Altaheri et al.~\cite{altaheri2023}. 
Additionally, predictive methods for multivariate time series have been systematically summarized by Lara et al.~\cite{lara2021} and Liang et al.~\cite{liang2024}. 
Furthermore, deep state space models (Deep SSMs) have recently emerged as a promising architecture for sequence modeling, with advances documented by Patro et al.~\cite{patro2024} and Wang et al.~\cite{wang2024}.

\paragraph{Deep learning-based EEG decoding.}
Deep learning-based methods for decoding EEG signals have been extensively investigated~\citep{EEGPT, EEGConformer}, and they demonstrate considerable potential for enhancing communication and rehabilitation in patients with physical paralysis~\citep{abibullaev2023, altaheri2023}. 
Early attempts rely on convolutional architectures~\citep{EEGNet, FBCNet}.
Nonetheless, the limited receptive field of convolutional layers has motivated the exploration of Transformer-based approaches~\citep{EEGConformer, Medformer} to model temporal dependencies in EEG signals.
Medformer~\citep{Medformer}, for instance, segments input signals into patches of varying temporal lengths and then feeds these patches into a Transformer. 
These Transformer-based approaches typically employ temporal patching strategies prior to Transformer processing to address computational complexity in long-sequence modeling. 
However, this preprocessing step may disrupt fine-grained temporal dependencies in EEG signals. 

\paragraph{Multivariate time-series forecasting models.}
Multivariate time-series forecasting has found broad applications across diverse domains, ranging from medical signal prediction (e.g., electromyography and electrocardiography analysis~\citep{zhang2022, METS}) as well as other applications~\citep{bi2023, huang2024}. 
Notably, Transformer-based approaches are widely studied due to their demonstrated capability for long-range sequence modeling (e.g.,~\citep{Informer, CrossFormer}). 
For example, Informer~\citep{Informer} proposed an efficient forecasting method based on a ProbSparse self-attention mechanism and a generative decoder. 
Other notable examples include PatchTST~\citep{PatchTST}, which divides time-series signals into smaller patches for Transformer input, and Crossformer~\citep{CrossFormer}, which introduces a two-stage attention mechanism to capture inter-variable and temporal dependencies separately.
In contrast, DLinear~\citep{DLinear} achieves performance comparable to Transformer-based approaches by utilizing a simple yet effective MLP-based method that decomposes time-series into seasonal and trend components. 
Drawing on these findings, iTransformer~\citep{iTransformer} proposes a strategy in which the matrix operations in the Transformer are transposed, thereby capturing inter-variable dependencies through the Attention mechanism while modeling temporal dependencies with an MLP.

\paragraph{Deep state space models.}
Although the Transformer architecture has been widely adopted across various domains~\citep{wav2vec, ViT}, the quadratic computational complexity $O(N^2)$ of its attention mechanism with respect to sequence length $N$ introduces fundamental scalability bottlenecks.
This limitation has motivated extensive research into efficient alternative architectures~\citep{Combiner, FlattenTransformer}.
Among these alternatives, Deep SSMs~\citep{S5, Mamba} have emerged as a promising framework for efficiently capturing long-range dependencies. 
In line with this trend, EEG classification models~\citep{EEG-SSM, EEGMamba} have also leveraged Deep SSMs to capture temporal dependencies in signals.
However, current models predominantly adopt Mamba~\citep{Mamba} as their foundational Deep SSM, leaving the optimal choice for the EEG classification task ambiguous. 
In contrast, our proposed method extends S5~\citep{S5}, a Deep SSM explicitly designed to capture inter-variable dependencies within the state space, to effectively address the intrinsic multivariate nature of EEG signals.

\section{Preliminaries}
\paragraph{Deep state space models}
Recent advancements in Deep SSMs~\citep{S5, Mamba2} have demonstrated their remarkable advantages over predominant architectures (including Transformer~\citep{Transformer}) across various sequence modeling tasks. 
Inspired by classical SSM~\citep{KalmanFilter}, Deep SSMs establish a principled framework in which input signals $\mathbf{x}(t)\in\mathbb{R}^P$ are mapped to output signals $\mathbf{y}(t)\in\mathbb{R}^P$ via latent states $\mathbf{h}(t)\in\mathbb{R}^Q$, as follows:

\begin{align}
    \label{eq:mimo_ssm}
    \frac{d\mathbf{h}(t)}{dt} = \mathbf{A}\mathbf{h}(t) + \mathbf{B}\mathbf{x}(t) ,\qquad \mathbf{y}(t) = \mathbf{C}\mathbf{h}(t) + \mathbf{D}\mathbf{x}(t),
\end{align}

where $P$ and $Q$ denote the number of variables for the input/output signals and the latent states, respectively. 
Moreover, $\mathbf{A}\in\mathbb{R}^{Q\times Q}$ represents the state matrix, while $\mathbf{B}\in\mathbb{R}^{Q\times P}$, $\mathbf{C}\in\mathbb{R}^{P\times Q}$, and $\mathbf{D}\in\mathbb{R}^{P\times P}$ denote the projection matrices.
Notably, the Deep SSM variant S5~\citep{S5} has demonstrated remarkable effectiveness in capturing sequential relationships for continuous signals. 
Here, we detail the processing steps employed in S5.

In S5, the HiPPO-N matrix~\citep{S4} is adopted as $\mathbf{A}$ to effectively capture long-range dependencies in sequential signals. 
Since the HiPPO-N matrix is real symmetric, it can be diagonalized as $\mathbf{A}=\mathbf{V\Lambda V}^{-1}$, thereby transforming Equation~(\ref{eq:mimo_ssm}) into the following form:
\begin{align}
    \label{eq:mimo_ssm2}
    \frac{d\tilde{\mathbf{h}}(t)}{dt} = \mathbf{\Lambda}\mathbf{\tilde{h}}(t) + \mathbf{\tilde{B}}\mathbf{x}(t),\qquad\mathbf{y}(t)=\mathbf{\tilde{C}}\mathbf{\tilde{h}}(t) + \mathbf{D}\mathbf{x}(t),
\end{align}

where $\tilde{\mathbf{h}}(t)=\mathbf{V}^{-1}\mathbf{h}(t)$, $\tilde{\mathbf{B}}=\mathbf{V}^{-1}\mathbf{B}$, and $\tilde{\mathbf{C}}=\mathbf{CV}$. 
Moreover, by introducing a timescale parameter $\mathbf{\Delta}\in\mathbb{R}_+$, Equation~(\ref{eq:mimo_ssm2}) is discretized using the zero-order hold (ZOH) method~\citep{zero-hold} as follows:
\begin{align}
    \label{eq:mimo_ssm3}
    \mathbf{\tilde{h}}_t = \mathbf{\bar{\Lambda}}\mathbf{\tilde{h}}_{t-1} + \mathbf{\bar{B}}\mathbf{x}_t,\qquad\mathbf{y}_t=\mathbf{\bar{C}}\mathbf{\tilde{h}}_t + \mathbf{\bar{D}}\mathbf{x}_t,
\end{align}
where $\bar{\mathbf{\Lambda}}=\mathrm{exp}(\mathbf{\Lambda}\mathbf{\Delta})$, $\bar{\mathbf{B}}=\mathbf{\Lambda}^{-1}\left(\bar{\mathbf{\Lambda}}-\mathbf{I}\right)\tilde{\mathbf{B}}$, $\bar{\mathbf{C}}=\tilde{\mathbf{C}}$, $\bar{\mathbf{D}}=\mathbf{D}$.
In practice, $\mathbf{\Delta}\in\mathbb{R}^Q$ is used for the timescale parameter, and $\mathbf{D}$ is restricted as a diagonal matrix.
Under these conditions, the learnable parameters consist of $\mathrm{diag}(\mathbf{\Lambda})$, $\tilde{\mathbf{B}}$, $\tilde{\mathbf{C}}$, $\mathrm{diag}(\mathbf{D})$, and $\mathbf{\Delta}$. 
Furthermore, S5 achieves efficient modeling of Equation~(\ref{eq:mimo_ssm3}) through the introduction of parallel scanning.

\begin{figure}
    \centering
    \includegraphics[width=\textwidth]{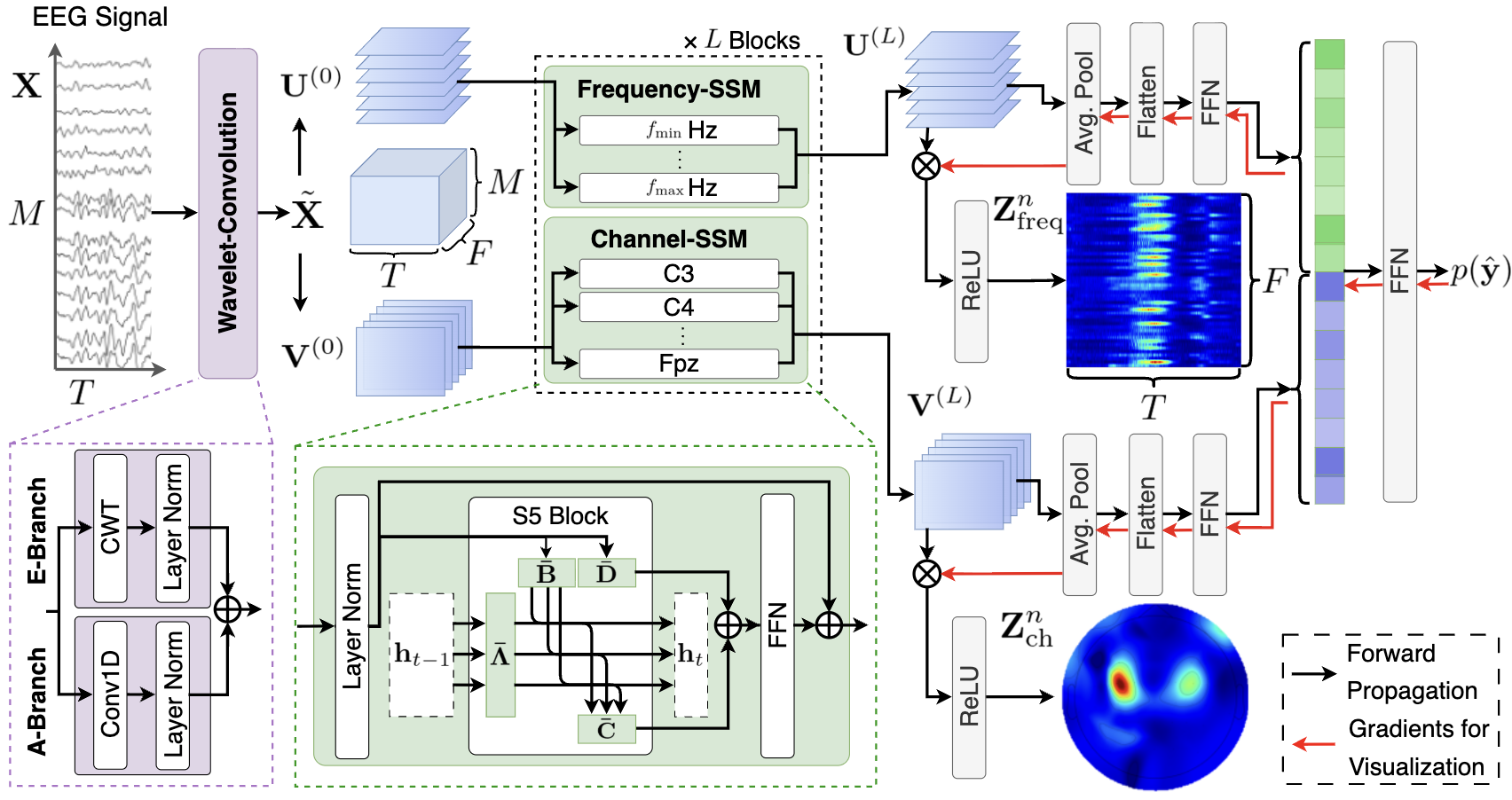}
    \vspace{-2em}
    \caption{Overview of the proposed Cortical-SSM. Given an EEG signal, the Wavelet-Convolution module extracts interpretable frequency features by combining deterministic and adaptive features (left). The Frequency-SSM and Channel-SSM then model spatio-temporal features for each frequency component and temporal-frequency features for each electrode, respectively (middle). Finally, the resulting features $\mathbf{U}^{(L)}$ and $\mathbf{V}^{(L)}$ are fused for action prediction, while providing visual explanations in spatio-temporal and temporal-frequency domains, respectively (right).
    }
    \label{fig:model}
    \vspace{-0.5em}
\end{figure}

\section{Method}
\label{section:method}
\subsection{Cortical-SSM}

In this study, we address a classification task for EEG signals recorded while subjects engage in motor imagery (MI) tasks.
In this task, it is desirable to accurately classify the corresponding EEG signals recorded while subjects engage in MI involving $N$ types of movement.
The input is EEG signals recorded while the subjects performed MI, and the output is a predicted probability corresponding to the MI. 

Unlike previous studies~\cite{EEG-SSM, EEGMamba}, our method jointly models spatio-temporal features within individual frequency components and temporal-frequency features within individual electrodes. This parallel design yields a comprehensive multi-domain representation for EEG classification. Because the architecture is not specific to MI, it can also be applied to decoding other endogenous and exogenous stimuli from EEG signals.

Figure~\ref{fig:model} shows the overall architecture of our proposed Cortical-SSM.
The proposed method consists of three main modules, Wavelet-Convolution, Frequency-SSM, and Channel-SSM.
The input $\mathbf{X}\in\mathbb{R}^{M\times T}$ to our model is the recorded EEG signal, where $M$ and $T$ denote the number of electrodes and the sequence length, respectively.

\subsection{Wavelet-Convolution}
In the Wavelet-Convolution module, we integrate deterministically derived frequency components with learned frequency features.
Previous models for EEG classification widely employed one-dimensional convolutional layers to learn frequency features through their kernels~\citep{EEGNet, EEGConformer}.
Despite the promising results achieved by these models, their black-box nature poses challenges for direct interpretability and impedes frequency domain analysis of the extracted features.
In contrast, the conventional short-time Fourier transform (STFT) and wavelet transform extract frequency components in a deterministic manner, thereby yielding inherently interpretable features. 
However, these fixed representations constrain the model's performance (see Subsection~\ref{section:ablation} for details).

To address this trade-off between learnability and interpretability, we propose a module integrating deterministically derived frequency components from an Explanation Branch (E-Branch) with trainable frequency features from an Adaptation Branch (A-Branch). 
This architecture maintains both the learnability of these representations and the extraction of non-black-box features.
Specifically, we employ the continuous wavelet transform (CWT) for the E-Branch (first term of Equation~(\ref{eq:wavelet-conv})), and a one-dimensional convolutional layer for the A-Branch (second term of Equation~(\ref{eq:wavelet-conv})).
Unlike existing methods that integrate the wavelet transform within convolutional layers~\citep{WaveletPooling, MWCNN}, our module processes CWT and 1D convolutional features in parallel, thereby enabling effective feature fusion.

Let $\mathbf{x}_m\in\mathbb{R}^{T}$ denote the EEG signal recorded from the $m$-th electrode ($m=1\ldots M$). The module's output $\mathbf{\tilde{x}}_m\in\mathbb{R}^{F\times T}$ for each $\mathbf{x}_m$ is formulated as shown below, yielding $\mathbf{\tilde{X}} = \left\{\mathbf{\tilde{x}}_m\ \middle|\ m=1\ldots M\right\}\in\mathbb{R}^{M\times F\times T}$:
\begin{align}
    \tilde{\mathbf{x}}_m =  \frac{1}{2}\mathrm{LayerNorm}\left(\mathrm{CWT}\left(\mathbf{x}_m\right)\right) + \frac{1}{2}\mathrm{LayerNorm}\left(\mathrm{Conv1D}\left(\mathbf{x}_m\right)\right),
    \label{eq:wavelet-conv}
\end{align}
where $\mathrm{Conv1D}(\cdot)$, $\mathrm{CWT}(\cdot)$, and $\mathrm{LayerNorm}(\cdot)$ represent one-dimensional convolutional layers, CWT, and layer normalization~\citep{LayerNorm}, respectively.
Additionally, $F$ represents the dimension in the frequency domain.
Details of the processing are described below.

First, in $\mathrm{Conv1D}(\cdot)$, the kernel length $K$ is set to $K=f_\mathrm{sample}/2$ following EEGNet~\citep{EEGNet}, and frequency features are extracted. 
Here, $f_\mathrm{sample}$ represents the sampling frequency.

In $\mathrm{CWT}(\cdot)$, a filter bank $\mathbf{\Psi}\in\mathbb{R}^{F\times T}$ is constructed to extract features for individual frequency components.
For the extraction of frequency components, the Morlet wavelet~\citep{torrence1998} is employed as the standard mother wavelet.
Let $\bm{\psi}_f=\left[\psi_{f,-\frac{T}{2}},\psi_{f,-\frac{T}{2}+1},\ \ldots,\ \psi_{f,\frac{T}{2}}\right]\in\mathbb{R}^T$ denote the Morlet wavelet used to extract a frequency component $f$.
$\psi_{f,t}$ is defined as follows:
\begin{align}
    \psi_{f,t}=\sqrt{\frac{1}{s}}\pi^{-\frac{1}{4}}e^{i\omega_{0}\frac{t}{s}}e^{-\frac{1}{2}{\left(\frac{t}{s}\right)}^2},\ s = \frac{\omega_0 f_\mathrm{sample}}{2\pi f},
\end{align}
where $\omega_0$ and $s$ denote the center frequency and scaling factor, respectively. 
Based on the above, $\mathbf{\Psi}$ is expressed by the following equation:
\begin{align}
  \mathbf{\Psi}=\left\{\psi_f\ \middle|\ f=f_{\mathrm{min}}+\alpha (f_{\mathrm{max}} - f_{\mathrm{min}})/F,\ \alpha=1\ldots F\right\},
\end{align}
where $f_\mathrm{min}$ and $f_\mathrm{max}$ denote the minimum and maximum $f$ targeted by the CWT, respectively. 

Finally, we employ $\mathrm{LayerNorm}(\cdot)$ to normalize the features obtained from $\mathrm{Conv1D}(\cdot)$ and $\mathrm{CWT}(\cdot)$.
The learnable affine parameters in $\mathrm{LayerNorm}(\cdot)$ adaptively modulate the feature distributions of both branches, effectively acting as an implicit gating mechanism. 
This enables channel-wise and frequency-wise adaptive weighting prior to fusion, rather than enforcing a fixed combination. Furthermore, $\mathrm{LayerNorm}(\cdot)$ is specifically applied along the temporal dimension, as explained below.
Normalization techniques widely applied in deep learning include batch normalization~\citep{BatchNorm}, group normalization~\citep{GroupNorm}, and layer normalization across dimensions at a given time step. 
These methods normalize features across variables. 
However, in multivariate time-series, when an arbitrary event occurs across variables and its effects appear at different time steps in the sequence, normalization across variables is known to introduce mutual noise~\citep{NonStationaryTransformer, RevIn}. 
Similarly, in EEG classification tasks, when internal stimuli derived from motor activity are recorded as signal sources via electrodes, these stimuli influence different temporal points within each electrode’s signal.
Therefore, we adopt layer normalization along the temporal dimension to address these issues. 

\subsection{Frequency-SSM}
\label{section:freuency-ssm}
The Frequency-SSM module independently captures spatio-temporal feature interactions within each individual frequency component.
Previous EEG studies~\citep{pfurtscheller2001, miller2007} have reported that MI tasks elicit frequency-specific power variations localized to functionally relevant cortical regions. 
Motivated by these neurophysiological findings, Frequency-SSM explicitly models the spatio-temporal dependencies for each frequency component in an independent manner. 
This design effectively tracks the power variations of those frequency bands associated with MI.
The proposed module comprises $L$ hierarchically organized blocks, where each block includes layer normalization, a feed-forward network, and a Deep SSM.
In the following, we provide a detailed description of the processing steps applied to the input $\mathbf{U}^{(l)}$ within the $l$-th block ($l=0,\ldots ,L$), where $\mathbf{U}^{(0)}=\tilde{\mathbf{X}}$.

We first normalize $\mathbf{U}^{(l)}$ along the temporal dimension for each frequency component $f\in\{1,\ldots,F\}$:
\begin{align}
    \tilde{\mathbf{u}}_f^{(l)} = \mathrm{LayerNorm}(\mathbf{u}_f^{(l)}).
\end{align}
Next, we capture the temporal dependencies in $\tilde{\mathbf{u}}_f^{(l)}$. 
In EEG classification tasks, Transformer-based approaches are widely employed to model temporal dependencies~\citep{Medformer,EEGPT}.
However, for long-sequence modeling, Deep SSMs have demonstrated superior performance to Transformer~\citep{S4, Mamba}. 
Therefore, our module incorporates a Deep SSM to capture temporal dependencies in EEG signals.
Deep SSMs can be broadly categorized into time-invariant~\citep{SaShiMi,S5} and time-varying~\citep{Mamba, Mamba2} systems. 
Although time-varying Deep SSMs are prevalent in EEG classification~\citep{EEG-SSM,EEGMamba}, prior work~\citep{Mamba} suggests that their inherent selection mechanisms can be detrimental for continuous signals, and they have exhibited inferior performance in certain speech synthesis tasks.
Therefore, we opt for time-invariant Deep SSMs. 
Furthermore, Deep SSMs can be classified by their input-output configuration into Single-Input Single-Output (SISO)~\citep{S4,Mamba2} or Multi-Input Multi-Output (MIMO)~\citep{S5, S7}. 
Given the multi-electrode nature of EEG signals, a MIMO configuration that preserves inter-variable dependencies within the state space is deemed appropriate. 
For these reasons, we extend S5~\citep{S5} as the Deep SSM with a time-invariant and MIMO configuration. 

By defining the operation of S5 as $\mathrm{SSM}(\cdot)$, the feature $\mathbf{u}_f^{(l+1)}$ that captures temporal dependencies in $\tilde{\mathbf{u}}_f^{(l)}$ is obtained as the following equation:
\begin{align}
    \mathbf{u}_f^{(l+1)} = \mathrm{FFN}\left(\mathrm{SSM}\left(\tilde{\mathbf{u}}_f^{(l)}\right)\right) + \tilde{\mathbf{u}}_f^{(l)},
\end{align}
where $\mathrm{FFN}(\cdot)$ denotes the feed-forward network.
Subsequently, the output $\mathbf{{U}}^{(l+1)} = \left[ \mathbf{{u}}^{(l+1)}_1,\ \mathbf{{u}}^{(l+1)}_2,\ \ldots\ \mathbf{{u}}^{(l+1)}_F\right]\in\mathbb{R}^{M\times F\times T}$ of the module at the $l$-th block is obtained, capturing temporal dependencies independently for each frequency component.

\subsection{Channel-SSM}
\label{section:channel-ssm}
In the Channel-SSM, temporal-frequency features are extracted independently for each electrode. 
By explicitly modeling electrode-specific dependencies of temporal-frequency features, we capture localized variations in signal intensity associated with MI. 
This module comprises $L$ stacked blocks, each incorporating layer normalization, a feed-forward network, and Deep SSM layers hierarchically.
The input $\mathbf{V}^{(l)}$ within the $l$-th block ($l=0,\ldots ,L$) is modeled as follows, where $\mathbf{V}^{(0)}=\tilde{\mathbf{X}}$:
\begin{align}
    \mathbf{v}_m^{(l+1)} = \mathrm{FFN}\left(\mathrm{SSM}\left(\mathrm{LayerNorm}\left({\mathbf{v}}_m^{(l)}\right)\right)\right) + \mathrm{LayerNorm}\left({\mathbf{v}}_m^{(l)}\right),
\end{align}
yielding the block’s final output  $\mathbf{{V}}^{(l+1)} = \left[ \mathbf{{v}}^{(l+1)}_1,\ \mathbf{{v}}^{(l+1)}_2,\ \ldots\ \mathbf{{v}}^{(l+1)}_M\right]\in\mathbb{R}^{M\times F\times T}$.

Finally, the outputs $\mathbf{U}^{(L)}$ and $\mathbf{V}^{(L)}$ from Frequency-SSM and Channel-SSM, respectively, are integrated as follows, yielding the predicted probability $p(\hat{\mathbf{y}})\in\mathbb{R}^{N}$ of action corresponding to $\mathbf{X}$:
\begin{align}
 p(\mathbf{\hat{y}}) = \mathrm{FFN}\left(\left[\mathrm{AvgPool}\left(\mathbf{U}^{(L)}\right);\mathrm{AvgPool}\left(\mathbf{V}^{(L)}\right)\right]\right),
\end{align} 
where $N$ represents the number of action types.
Moreover, $\mathrm{AvgPool}(\cdot)$ denotes the average pooling layer, which aggregates the input features along the temporal dimension following the approach proposed in S4~\citep{S4}.
Furthermore, we use the cross-entropy loss as the loss function.

\subsection{Visual Explanations}
\label{section:inerpretability}
In our proposed method, we generate visual explanations in both the spatio-temporal and temporal-frequency domains through the following procedures. 
Our approach extends Grad-CAM~\citep{Grad-CAM} to generate visual explanations tailored for time-series signals.
Grad-CAM is formulated as follows, generating a feature map $\mathbf{Z}^n\in\mathbb{R}^{I\times J}$ that represents the explanation for the $n$-th output class from the gradient of the loss with respect to $\hat{y}_n$:
\begin{eqnarray}
   \alpha_r= \frac{1}{IJ}\sum_{i\in I}\sum_{j\in J}\frac{\partial \hat{y}^{(n)}}{\partial o_{r,i,j}}\ ,\qquad
   \mathbf{Z}^{(n)}=\mathrm{ReLU}\left(\sum_{r\in R}\alpha_r \mathbf{o}_r\right),
\end{eqnarray}
where $\hat{y}_n$ denotes the $n$-th element of $\hat{\mathbf{y}}$.
$I$, $J$, and $R$ denote the number of vertical and horizontal pixels, and the number of dimensions of the feature map, respectively.
Furthermore, $\mathbf{o}_r\in\mathbb{R}^{I\times J}$ and $o_{r,i,j}\in\mathbb{R}$ denote the feature map of the $r$-th dimension and the feature indexed by position $\left(i,j\right)$ within $\mathbf{o}_r$, respectively. 
In this context, $\mathbf{o}_r$ is required to retain the spatial relationships of the input.

Next, we describe our method for generating visual explanations in the spatio-temporal domain. 
While Grad-CAM operates on 2D feature maps $\mathbf{o}_r$ and performs weighting across $\left\{\mathbf{o}_r\right\}_{r=1}^R$, our method deals with time-series signals and therefore performs weighting across 1D time-series vectors. 
Here, we generate visual explanations using $\mathbf{V}^{(L)}$, the feature closest to the output layer, which preserves the spatio-temporal relationships of the input signal. 
As detailed in Subsection \ref{section:channel-ssm}, the features within $\mathbf{V}^{(L)}$ are processed independently for each electrode.
Therefore, we generate a separate visual explanation for each electrode by weighting across $\left\{\mathbf{v}_{f,m}^{(L)}\right\}_{f=1}^F$. 
This yields the feature map $\mathbf{Z}_\mathrm{ch}^{n}\in\mathbb{R}^{M\times T}$, representing the spatio-temporal visual explanation for the $n$-th output class, obtained as follows:
\begin{eqnarray}
\label{eq:visualization}
  \alpha^n_{f,m}&=&\frac{1}{T}\sum_{t\in T}\frac{\partial \hat{y}_n}{\partial {v}^{(L)}_{f,m,t}},\\
\label{eq:visualization２}
  \mathbf{Z}^n_{\mathrm{ch}}&=&\left\{\mathrm{ReLU}\left(\sum_{f\in F}\alpha^n_{f,m}\mathbf{v}^{(L)}_{f,m}\right)\ \middle|\ m=1\ldots M\right\}.
\end{eqnarray}

Finally, we describe our method for generating visual explanations in the temporal-frequency domain. 
Here, we generate visual explanations using $\mathbf{U}^{(L)}$, the feature closest to the output layer, which preserves the temporal-frequency relationships of the input signal. 
As detailed in Subsection \ref{section:freuency-ssm}, in contrast to $\mathbf{V}^{(L)}$, the features within $\mathbf{U}^{(L)}$ are processed independently for each frequency component.
Therefore, we generate a separate visual explanation for each frequency component by weighting across $\left\{\mathbf{u}_{f,m}^{(L)}\right\}_{m=1}^M$. 
This yields the feature map $\mathbf{Z}_\mathrm{freq}^{n}\in\mathbb{R}^{F\times T}$, representing the temporal-frequency visual explanation for the $n$-th output class, as in Equations~(\ref{eq:visualization}) and (\ref{eq:visualization２}).

\begin{table}[t]
    \centering
    \caption{
        Implementation details of the proposed method. The table summarizes the hyperparameters of the EEG input configuration, Wavelet-Convolution module, Frequency-SSM and Channel-SSM modules, and the training settings used in all experiments. Symbols denote the corresponding notations used throughout the manuscript.
    }
    \label{tab:hyperparams}
    \renewcommand{\arraystretch}{1}
    \resizebox{0.8\textwidth}{!}{
    \begin{tabular}{l c c c}
      \toprule
      Taxonomy & Hyperparameter& Symbol& Setting\\
      \midrule
      \multirow{3}{*}{\makecell[l]{EEG\\sample}}& Channels& $M$& \makecell{62 (OpenBMI)\\ 64 (Stieger2021)}\\
      & Sampling rate [Hz]& $f_{\rm{sample}}$& 250\\
      & Segment duration [s]& --& 4\\
      & Time points& $T$& $250\times4=1000$\\
      \midrule
      \multirow{5}{*}{\makecell[l]{Wavelet-\\Convolution}}& Kernel Size& $K$& 250/2=125\\
      & Frequency dimension& $F$& 50\\
      & Minimum frequency [Hz] &$f_{\rm{min}}$& $1$\\
      & Maximum frequency [Hz] &$f_{\rm{max}}$& $100$\\
      & Central frequency [Hz] & $\omega_0$& 6\\
      \midrule

      \multirow{4}{*}{\makecell[l]{Frequency-SSM\\Channel-SSM}}& Block number& $L$& 2\\
      & State size& $Q$& \makecell[c]{64/2=32 (Channel-SSM)\\ 50/2=25 (Frequency-SSM)}\\
      & Feed-forward dimension& --& 256\\
            
      \midrule
      \multirow{9}{*}{Training}&Epoch& --& 100\\
      &Patience epoch& --& 5\\
      &Batch size& --& 8\\
      &Optimizer& --& AdamW\\
      &Adam $\beta$& --&{(0.9, 0.999)}\\
      &Learning rate& --& $1.0\times10^{-4}$\\
      &Weight decay& --& $1.0\times10^{-4}$\\
      &Scheduler& --& CosineAnnealingLR\\
      &Cosine cycle epochs& --& 100\\
      \bottomrule
    \end{tabular}
    }
    \vspace{-1em}
\end{table}

\begin{table}[t]
    \centering
    \caption{
        Comparison of model complexity and computational efficiency. We report the number of trainable parameters and inference time (Inf. Time) for the proposed method and all baselines. The proposed method was comparable to the baselines in terms of both model complexity and computational efficiency.
    }
    \label{tab:model_param_comp}
    \renewcommand{\arraystretch}{1}
    \resizebox{0.6\textwidth}{!}{
    \begin{tabular}{l l c c}
      \toprule
      Taxonomy & Model& \makecell{Param$\downarrow$ \\{[M]}}&  \makecell{Inf. Time$\downarrow$ \\{[ms]}}\\
      \midrule
      \multirow{12}{*}{\makecell[l]{General\\Time-Series\\Model}}& Informer~\cite{Informer}& 0.68& 1.48\\
      &Autoformer~\cite{Autoformer} &0.67& 1.70\\
      &FEDformer~\cite{FEDformer}&0.67 &21.30\\
      &Crossformer~\cite{CrossFormer}&2.02 &1.83\\
      &DLinear~\cite{DLinear}&2.13 &0.11\\
      &TimesNet~\cite{TimesNet}&1.21 &5.73\\
      &PatchTST~\cite{PatchTST}&1.24 &0.83\\
      &TimeMixer~\cite{TimeMixer}&6.46 &1.50\\
      &iTransformer~\cite{iTransformer}&0.74 &0.69\\
      &UniTS~\cite{UniTS}&2.32 &2.09\\
      &TimeMachine~\cite{TimeMachine}& 0.90& 0.73\\
      &S-Mamba~\cite{S-Mamba}& 2.16& 1.28\\
      \midrule
      \multirow{7}{*}{\makecell[l]{EEG\\Decoding\\Model}}& Shallow ConvNet~\cite{DeepConvNet}& 0.10& 0.52\\
      &Deep ConvNet~\cite{DeepConvNet}& 0.30& 1.40\\
      &EEGNet~\cite{EEGNet}& 0.02& 0.17 \\
      &TSception~\cite{TSCeption}& 0.03& 0.35\\
      &EEG Conformer~\cite{EEGConformer} &0.85& 1.34 \\
      &Medformer~\cite{Medformer} &2.32 & 4.67\\
      &Cortical-SSM (Ours)& 0.83 & 2.20\\
      \bottomrule
    \end{tabular}
    }
    \vspace{-1em}
\end{table}

\section{Experiments}
\label{section:exp_setup}
\subsection{Datasets and Data Pre-Processing}
In our experiments, we employed two publicly available MI EEG datasets, OpenBMI~\citep{OpenBMI} and Stieger2021~\citep{Stieger}. These datasets were chosen because previous studies have demonstrated substantial domain shifts across subjects in MI EEG recordings~\citep{Cho, Kaya}, making them suitable for evaluating the generalizability and robustness of decoding models. 
Both datasets include multiple sessions collected from more than fifty participants, enabling systematic subject-independent evaluation. 
Accordingly, we adopted a cross-subject $k$-fold cross-validation scheme ($k=8$).

For preprocessing, we followed the minimal pipeline described in~\cite{delorme2023}. 
EEG signals were downsampled to 250~Hz and used directly as model inputs without explicit noise or artifact removal.
Training, validation, and test sets were constructed following the procedures described below and were used for model training, hyperparameter tuning, and performance evaluation, respectively. 
During training, we monitored classification accuracy on the validation set at each epoch, and model parameters achieving the highest validation accuracy were retained for final evaluation on the held-out test set.
Further details regarding each dataset are provided below.

\paragraph{OpenBMI.}
OpenBMI is a public EEG dataset containing recordings from 54 healthy participants engaging in a two-class MI task involving left- and right-hand grasping~\citep{OpenBMI}. The experimental protocol follows~\citep{pfurtscheller2001}, where visual cues indicate the movement to be imagined. Each subject performed two sessions consisting of 400 MI trials per session, resulting in 21{,}600 samples recorded at 1000\,Hz using 62 Ag–AgCl electrodes positioned according to the international 10--20 system~\citep{klem1999}. We extract the 4-second MI period for each trial.
For cross-validation ($k=8$), subjects are randomly partitioned into training (44 subjects), validation (5 subjects), and test (5 subjects), corresponding to 17{,}600, 2{,}000, and 2{,}000 samples, respectively.

\paragraph{Stieger2021.}
Stieger2021 contains EEG recordings collected from 62 participants performing MI-based cursor control tasks~\citep{Stieger}. In this work, we focus on the left/right (LR) task, which involves imagined left- and right-hand grasping. Cursor movement was controlled using alpha power extracted from electrodes C3 and C4, and visual cues indicating the intended movement direction were presented on a screen. 
Across 7--11 sessions, each participant contributed between 1{,}050 and 1{,}650 MI trials, recorded using 64 electrodes (10--10 system) at 1000~Hz.
Following prior work, we focus on 202{,}950 samples from 41 participants who completed all sessions, and use 4-second windows centered around MI onset.
Subject-independent $k$-fold cross-validation was conducted with splits of 33, 4, and 4 subjects for training, validation, and testing, yielding 54{,}450, 6{,}600, and 6{,}600 samples, respectively.

\subsection{Baseline Methods}

This task involves (i) classification of EEG signals, and (ii) handling of multivariate time-series signals recorded from multiple electrodes. 
Accordingly, we selected baseline methods from the following perspectives:
\begin{itemize}
    \item[(i)] We chose baselines that have been successfully applied to EEG classification, similar to this task, including Shallow ConvNet~\citep{DeepConvNet}, Deep ConvNet~\citep{DeepConvNet}, EEGNet~\citep{EEGNet}, TSception~\citep{TSCeption}, EEG Conformer~\citep{EEGConformer}, and Medformer~\citep{Medformer}.
    \item[(ii)] Considering the multivariate nature of EEG signals, we selected baselines that have demonstrated effectiveness in multivariate time-series forecasting, such as Informer~\citep{Informer}, Autoformer~\citep{Autoformer}, FEDformer~\citep{FEDformer}, Crossformer~\citep{CrossFormer}, DLinear~\citep{DLinear}, TimesNet~\citep{TimesNet}, PatchTST~\citep{PatchTST}, TimeMixer~\citep{TimeMixer}, iTransformer~\citep{iTransformer}, S-Mamba~\citep{S-Mamba}, and TimeMachine~\citep{TimeMachine}.
\end{itemize}

\subsection{Implementation Details}
Table~\ref{tab:hyperparams} summarizes the experimental settings of the proposed method. 
The hyperparameters were fixed a priori such that the number of trainable parameters was comparable to that of the baselines.

Table~\ref{tab:model_param_comp} further reports the number of trainable parameters and the inference time for the proposed method and all baselines. 
The proposed model contained approximately 0.83 million parameters, with an inference time of 2.20 ms per sample, demonstrating computational cost comparable to the baselines.

All experiments were conducted on an NVIDIA GeForce RTX 4090 GPU (24 GB VRAM) and an Intel Core i9-13900KF CPU with 64 GB RAM. 
The training time per task for the proposed method was approximately 157 min on OpenBMI and 50 min on Stieger2021.

\section{Results}
\begin{table}[t]
    \centering
    \caption{
    Performance comparison on OpenBMI~\citep{OpenBMI} and Stieger2021~\cite{Stieger} datasets. \textbf{Bold} and \underline{underlined} values indicate the best and second-best performances, respectively.
    Results are reported as mean ± standard deviation over 8-fold cross-validation using accuracy, Macro-F1, AUROC, AUPRC, and Cohen’s Kappa.
    The proposed method achieves the best performance across all metrics on both datasets.
    }
    \label{tab:quantitative_main} 
    \resizebox{\textwidth}{!}{
    \begin{tabular}{c l c c c c c }
      \toprule
      \textbf{Dataset} &\textbf{Models} &\textbf{Accuracy~[\%]$~\uparrow$} &\textbf{Macro-F1~[\%]$~\uparrow$} &\textbf{AUROC~[\%]$~\uparrow$} &\textbf{AUPRC~[\%]$~\uparrow$} &\textbf{Kappa$~\uparrow$}\\
      \midrule
      \multirow{22}{*}{\shortstack{\textbf{OpenBMI}~\cite{OpenBMI}\\ (2 Classes)}}
      &Chance Performance &50.00 &50.00 &50.00 &50.00 &0.00\\
      \cmidrule(l{0mm}r{0mm}){2-7} 
      & \multicolumn{6}{l}{\textit{General Time-Series Models}}\\
      &Informer~\cite{Informer} &73.92\std5.19 &73.71\std5.25 &83.06\std5.79 &82.77\std5.75 &0.48\std0.10\\
      &Autoformer~\cite{Autoformer} &67.91\std4.84 &67.73\std4.94 &74.80\std5.79 &74.19\std6.04 &0.36\std0.10\\
      &FEDformer~\cite{FEDformer} &66.67\std4.83 &66.53\std4.85 &72.80\std6.71 &72.02\std7.22 &0.33\std0.10\\
      &Crossformer~\cite{CrossFormer} &71.85\std4.93 &71.74\std4.99 &80.05\std5.24 &79.82\std5.54 &0.44\std0.10\\
      &DLinear~\cite{DLinear} &70.46\std4.65 &70.34\std4.63 &70.56\std4.64 &64.84\std4.31 &0.41\std0.09\\
      &TimesNet~\cite{TimesNet} &73.19\std5.53 &73.10\std5.57 &80.58\std5.99 &79.95\std5.93 &0.46\std0.11\\
      &PatchTST~\cite{PatchTST} &76.80\std4.73 &76.73\std4.77 &84.94\std4.98 &83.87\std5.42 &0.54\std0.09\\
      &TimeMixer~\cite{TimeMixer} &54.08\std4.38 &48.91\std8.99 &54.62\std6.35 &54.03\std5.04 &0.08\std0.09\\
      &iTransformer~\cite{iTransformer} &69.83\std3.86 &69.75\std3.88 &78.04\std4.67 &78.44\std4.50 &0.40\std0.08\\
      &UniTS~\cite{UniTS}& 73.59\std4.63& 73.52\std4.67& 81.55\std5.52& 81.52\std5.56& 0.47\std0.09\\
      &TimeMachine~\cite{TimeMachine} &72.33\std4.36 &72.23\std4.37 &80.32\std5.29 &79.59\std6.17 &0.45\std0.09\\
      &S-Mamba~\cite{S-Mamba} &72.05\std4.20 &71.97\std4.22 &80.09\std4.97 &79.78\std5.26 &0.44\std0.08\\
      \cmidrule(l{0mm}r{0mm}){2-7}
      & \multicolumn{6}{l}{\textit{EEG Decoding Models}}\\
      &Shallow ConvNet~\cite{DeepConvNet} &75.21\std5.35 &75.15\std5.38 &82.20\std5.86 &80.93\std6.82 &0.50\std0.11\\
      &Deep ConvNet~\cite{DeepConvNet} &74.34\std6.68 &74.17\std6.84 &81.74\std7.27 &80.97\std7.61 &0.49\std0.13\\
      &EEGNet~\cite{EEGNet} &74.93\std4.48 &74.80\std4.52 &83.20\std5.11 &82.71\std5.26 &0.50\std0.09\\
      &TSception~\cite{TSCeption} &65.78\std5.77 &65.60\std5.87 &71.69\std7.84 &71.16\std8.20 &0.32\std0.12\\
      &EEG Conformer~\cite{EEGConformer} &\underline{77.90\std5.27} &\underline{77.78\std5.33} &\underline{85.81\std5.23} &\underline{85.36\std5.47} &\underline{0.56\std0.11}\\
      &Medformer~\cite{Medformer} &74.06\std3.35 &73.99\std3.37 &81.65\std4.05 &81.38\std4.38 &0.48\std0.07\\
      \cmidrule(l{0mm}r{0mm}){2-7} 
      &\multirow{2}{*}{} &\bf{81.62}\std\bf{\small5.20} 
      &\bf{81.56}\std\bf{\small5.24}
      &\bf{90.14}\std\bf{\small4.61}
      &\bf{90.17}\std\bf{\small4.47} 
      &\bf{0.63}\std\bf{\small0.10}\\
      &\multirow{-2}{*}{ 
      \textbf{Cortical-SSM (Ours)}} 
      & (+3.72) & (+3.78)
      & (+4.33) & (+4.81)
      & (+0.07)\\
\midrule
      \multirow{21}{*}{\shortstack{\bf{Stieger2021}~\cite{Stieger}\\ (2 Classes)}}
      & Chance Performance &50.00 &50.00 &50.00 &50.00 &0.00\\
      \cmidrule(l{0mm}r{0mm}){2-7} 
      & \multicolumn{6}{l}{\textit{General Time-Series Models}}\\
      & Informer~\cite{Informer} &77.90\std5.51 &77.85\std5.54 &86.83\std5.87 &87.05\std5.91 &0.56\std0.11\\
      & Autoformer~\cite{Autoformer} &72.02\std5.02 &71.96\std5.02 &79.75\std6.08 &79.22\std6.25 &0.44\std0.10\\
      & FEDformer~\cite{FEDformer} &67.48\std4.05 &67.16\std4.51 &74.46\std5.02 &73.36\std5.34 &0.35\std0.08\\
      & Crossformer~\cite{CrossFormer} &\underline{82.63\std4.81} &\underline{82.60\std4.83} &\underline{91.80\std4.03} &\underline{92.09\std3.89} &\underline{0.65\std0.10}\\
      & DLinear~\cite{DLinear} &63.66\std4.37 &63.65\std4.36 &63.70\std4.39 &58.94\std3.42 &0.27\std0.09\\
      & TimesNet~\cite{TimesNet} &77.43\std5.98 &77.41\std5.98 &86.37\std6.36 &86.54\std6.46 &0.55\std0.12\\
      & PatchTST~\cite{PatchTST} &79.67\std3.52 &79.64\std3.52 &88.85\std3.76 &88.47\std4.75 &0.59\std0.07\\
      & TimeMixer~\cite{TimeMixer} &53.97\std4.88 &46.91\std9.21 &54.29\std5.09 &53.33\std3.86 &0.08\std0.10\\
      & iTransformer~\cite{iTransformer} &78.67\std5.77 &78.62\std5.82 &87.85\std5.93 &88.08\std6.00 &0.57\std0.12\\
    & UniTS~\cite{UniTS} &82.59\std3.05 &82.58\std3.05 &91.21\std2.90 &91.38\std2.90 &\underline{0.65\std0.06}\\
      & TimeMachine~\cite{TimeMachine} &79.23\std5.83 &79.22\std5.83 &87.17\std5.93 &86.30\std6.44 &0.58\std0.12\\
      & S-Mamba~\cite{S-Mamba} &80.23\std5.70 &80.21\std5.71 &88.67\std5.73 &88.67\std5.73 &0.60\std0.11\\
      \cmidrule(l{0mm}r{0mm}){2-7}
      & \multicolumn{6}{l}{\textit{EEG Decoding Models}}\\
      & Shallow ConvNet~\cite{DeepConvNet} &52.22\std2.68 &42.95\std7.18 &64.44\std3.90 &60.60\std3.58 &0.04\std0.05\\
      & Deep ConvNet~\cite{DeepConvNet} &69.56\std2.97 &69.15\std2.86 &75.01\std6.09 &70.68\std6.67 &0.39\std0.06\\
      & EEGNet~\cite{EEGNet} &75.16\std5.92 &75.06\std5.93 &79.75\std6.95 &75.57\std7.31 &0.50\std0.12\\
      & TSception~\cite{TSCeption} &63.19\std6.38 &60.29\std9.98 &70.46\std3.00 &67.02\std2.99 &0.26\std0.13\\
      & EEG Conformer~\cite{EEGConformer} &72.53\std5.73 &72.42\std5.77 &79.90\std6.92 &77.89\std7.53 &0.45\std0.11\\
      & Medformer~\cite{Medformer} &78.67\std6.12 &78.61\std6.15 &88.06\std5.87 &88.47\std5.75 &0.57\std0.12\\
       \cmidrule(l{0mm}r{0mm}){2-7} 
      & \multirow{2}{*}{ } &\bf{87.12}\std\bf{\small4.33} 
      &\bf{87.11}\std\bf{\small4.33}
      &\bf{94.66}\std\bf{\small3.40}
      &\bf{94.82}\std\bf{\small3.32} 
      &\bf{0.74}\std\bf{\small0.09}\\
      &\multirow{-2}{*}{\bf{Cortical-SSM (Ours)}} 
      &(+4.49) &(+4.51)
      &(+1.86) &(+2.73)
      &(+0.09)\\
      \bottomrule
    \end{tabular}
    }
    \vspace{-1em}
\end{table}

\label{section:results}
\subsection{Quantitative Results}
\label{section:quantitative}
We conducted experiments to compare the performance of our model with baselines on the OpenBMI~\citep{OpenBMI} and Stieger2021~\citep{Stieger} datasets.
Quantitative results are summarized in Table~\ref{tab:quantitative_main}.
Values reported in the table represent the mean and standard deviation obtained across $k$-fold cross-validation ($k=8$).
For the evaluation metrics, we employed accuracy, Macro-F1, AUROC (macro-averaged), AUPRC (macro-averaged), and Cohen’s Kappa.
We used these metrics because they are standard for the classification of EEG signals.

\begin{figure}[t]
    \centering
    \includegraphics[width=\textwidth]{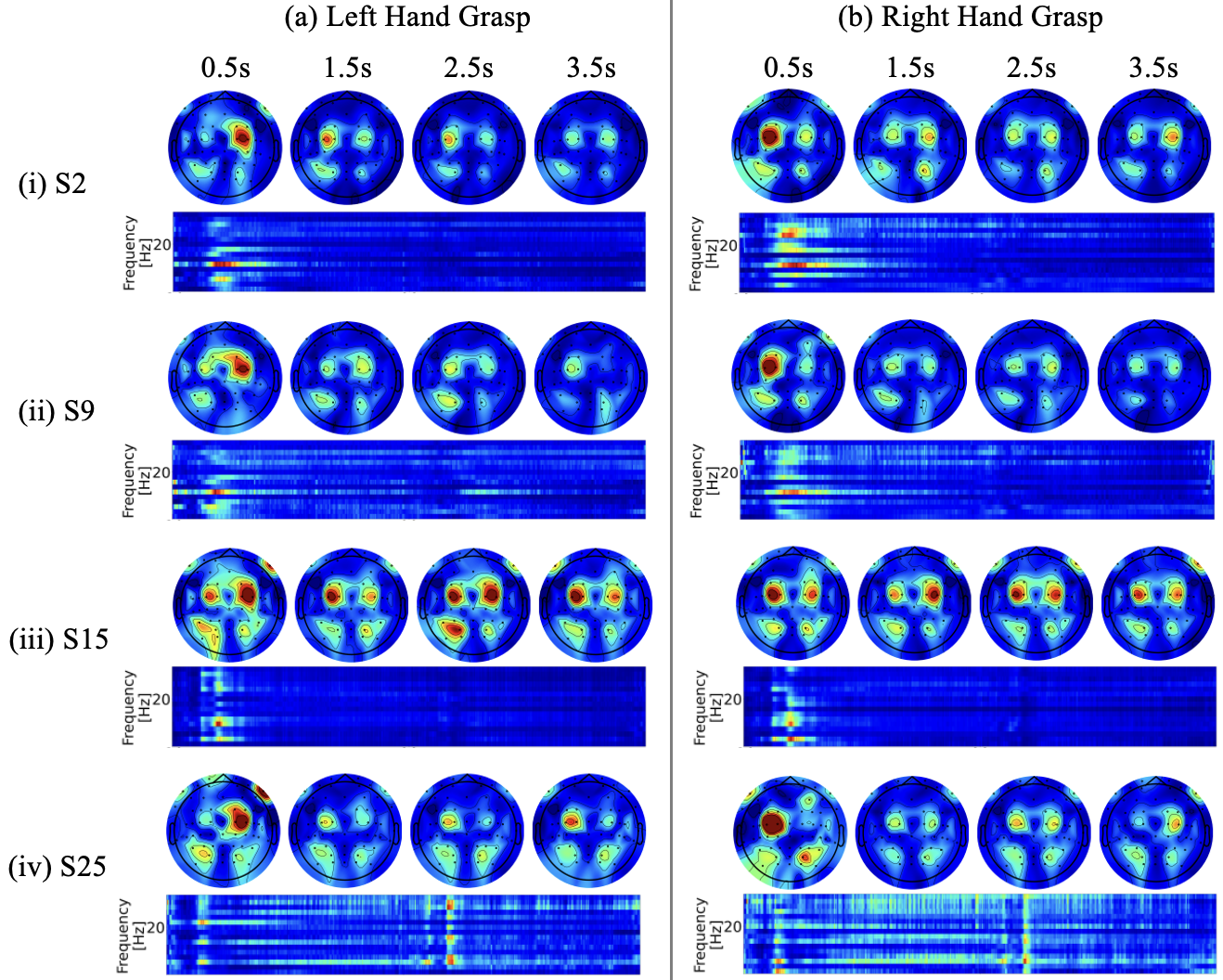} 
    \vspace{-1em}
    \caption{
        Sample-agnostic visual explanations of our proposed method on OpenBMI~\citep{OpenBMI}. Rows (i)–(iv) show results for Subjects S2, S9, S15, and S25, respectively. Columns (a) and (b) correspond to the left-hand grasp and right-hand grasp classes, respectively. Each visual explanation is presented at discrete time intervals within a 4-second MI trial.
    }
    \label{fig:avg_OpenBMI}
    \vspace{-0.5em}
\end{figure}


Table~\ref{tab:quantitative_main} presents the performance of our proposed method, which achieved  the highest accuracy, macro-F1, AUROC, AUPRC, and Cohen’s Kappa on OpenBMI, with values of 81.62\%, 81.56\%, 90.14\%, 90.17\%, and 0.63 points, respectively.
The proposed method outperformed the second-best model by 3.72, 3.78, 4.33, 4.81, and 0.07 points for these respective metrics. 
Moreover, the proposed method achieved state-of-the-art results on Stieger2021 with accuracy, macro-F1, AUROC, AUPRC, and Cohen’s Kappa of  87.12\%, 87.11\%, 94.66\%, 94.82\%, and 0.74 points, respectively.
Compared with the second-best model, our method exhibited  improvements of 4.49, 4.51, 1.86, 2.73, and 0.09 points for accuracy, macro-F1, AUROC, AUPRC, and Cohen’s Kappa, respectively.

\begin{figure}[t]
    \centering
    \includegraphics[width=\linewidth]{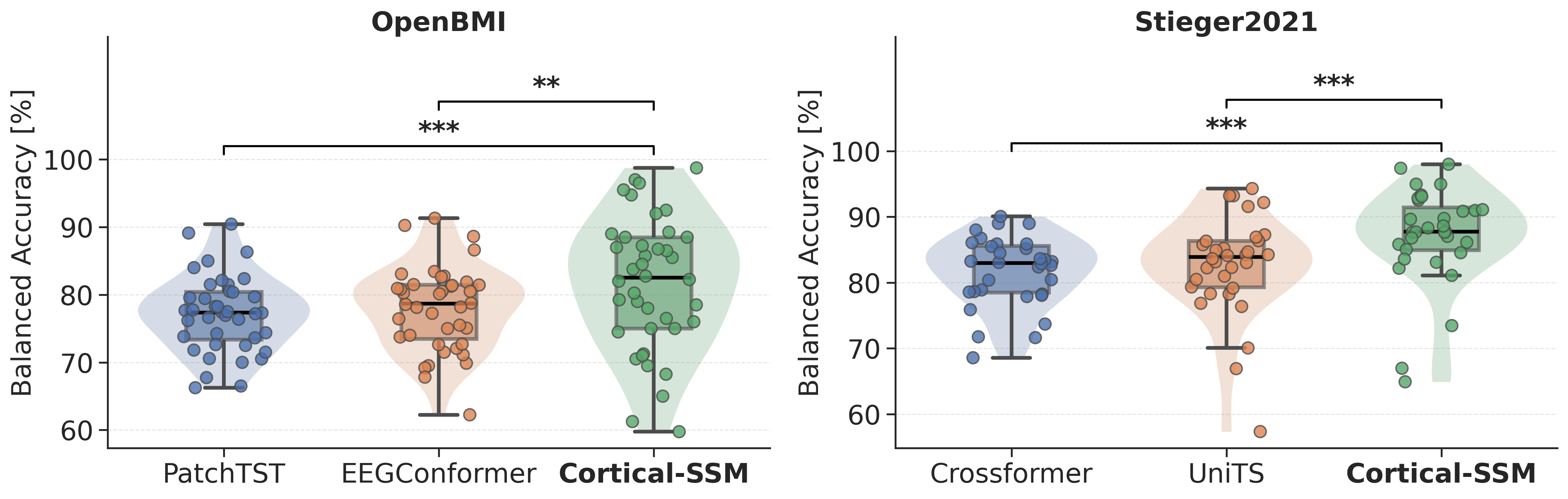}
    \vspace{-1em}
    \caption{
    Subject-wise comparison on OpenBMI~\cite{OpenBMI} and Stieger2021~\cite{Stieger}. Each point represents an individual subject. Violin and box plots summarize the distribution of performance across subjects. The proposed method outperformed the baselines for the majority of subjects, with statistically significant overall improvements (Wilcoxon signed-rank test, $p < 0.01$).
    }
    \label{fig:subj_quart}
\end{figure}

Furthermore, Figure~\ref{fig:subj_quart} reports subject-wise decoding performance on both benchmarks.
Specifically, for each dataset, we compare the proposed method with the top two baseline models identified in Table 1 in terms of classification accuracy.
Each point corresponds to an individual subject, while violin and box plots summarize the distribution across subjects, enabling a comprehensive assessment of inter-subject variability beyond aggregated statistics such as mean and standard deviation.

As shown in Figure~\ref{fig:subj_quart}, the proposed method consistently outperforms the baselines for the majority of subjects across both datasets.
To quantitatively assess these differences, we apply the Wilcoxon signed-rank test, a nonparametric test for paired samples, and observe statistically significant improvements across all evaluation metrics and both benchmarks ($p < 0.05$).
The use of a nonparametric test is justified by a normality analysis: Shapiro–Wilk tests conducted on fold-wise accuracy values reject the normality assumption ($p < 0.05$), likely due to subject-specific variability and domain shifts inherent in EEG signals~\citep{Cho, Kaya}.
These deviations from normality violate the assumptions of parametric tests such as the $t$-test, thereby necessitating a nonparametric alternative.
Overall, these results indicate that the observed performance gains are not driven by a small subset of subjects but are consistently achieved across the population.

\subsection{Interpretability}
We further examine the proposed method by generating visual explanations, following the procedure described in Section~\ref{section:inerpretability}.
Figure~\ref{fig:avg_OpenBMI} presents sample-agnostic visual explanations on the OpenBMI dataset~\citep{OpenBMI}.
Column (a) corresponds to left-hand grasp imagery, while Column (b) corresponds to right-hand grasp imagery.
Rows (i)–(iv) show representative subjects (S2, S9, S15, and S25, respectively).
For each subject and class, the upper panels depict spatio-temporal topographic visualizations at 0.5 s, 1.5 s, 2.5 s, and 3.5 s relative to MI onset, obtained by projecting $\mathbf{Z}_\mathrm{ch}^n$ (introduced in Section~\ref{section:inerpretability}) onto scalp maps aligned with the 10--20 system~\citep{klem1999}.
The lower panels illustrate the corresponding temporal–frequency visualizations, representing $\mathbf{Z}_\mathrm{freq}^n$, where 0 seconds denotes the onset of motor imagery.
Across subjects and both imagery classes, the temporal–frequency maps consistently highlight activity in the mu band (approximately 10 Hz), which is well known to be associated with motor imagery in EEG~\citep{pfurtscheller2006}.
Moreover, the topographic maps show strong attention around electrodes C3 and C4, which are located over the motor cortex and are neurophysiologically linked to hand motor control~\citep{pfurtscheller2006}.
These observations indicate that the proposed method reliably attends to neurophysiologically meaningful frequency bands and spatial regions, in a sample-agnostic manner.

\begin{figure}[t]
    \centering
    \includegraphics[width=0.85\linewidth]{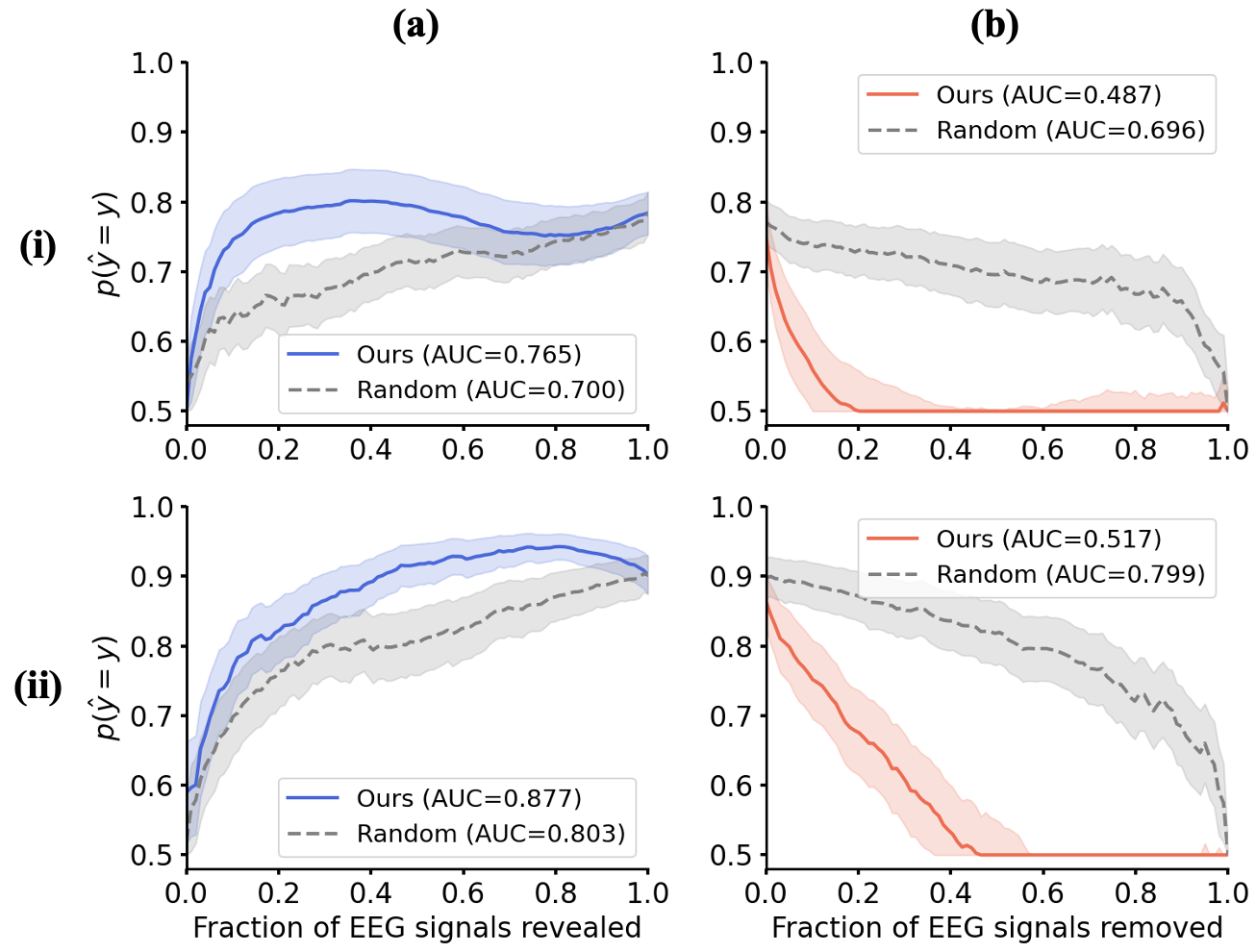}
    \caption{Quantitative evaluation of the faithfulness of the proposed explanations using insertion and deletion curves. Panels (a) and (b) show the insertion (blue) and deletion (red) curves, respectively, while panels (i) and (ii) correspond to results on 100 samples from OpenBMI and Stieger2021. Gray curves indicate random feature selection as a control, and the shaded regions represent ±3 standard deviations. The proposed method consistently outperformed the random baseline ($p<0.001$), achieving higher AUC for insertion and lower AUC for deletion, demonstrating that the identified features are critical for the model’s predictions.}
    \label{fig:ID_Score}
\end{figure}

Furthermore, to quantitatively assess the explanations, we evaluate their causal faithfulness using insertion and deletion curves~\cite{ID-Score}, as shown in Fig~\ref{fig:ID_Score}.
Insertion and deletion curves are standard metrics for evaluating the faithfulness of visual explanations. 
They quantify how the model’s prediction changes as regions identified as important by a saliency map are progressively inserted or removed.
Formally, given an input $\bm{X}$, we define a sequence of perturbed inputs $\{\bm{X}_k\}$ by inserting (or deleting) the top $k~\%$ of features ranked by the saliency map. 
For each $\bm{X}_k$, we compute the predicted probability of the ground-truth class, $p(\hat{y}_k=y)$. 
The insertion (or deletion) curve is then defined as the trajectory of $p(\hat{y}_k=y)$ as a function of $k$.
A faithful explanation is expected to yield a steep increase in the insertion curve and a steep decrease in the deletion curve for small $k$, indicating that the identified features are indeed critical for the model’s prediction.

In Fig.~\ref{fig:ID_Score}, panels (a) and (b) show the insertion (blue) and deletion (red) curves, respectively, while panels (i) and (ii) correspond to results on 100 samples from OpenBMI and Stieger2021. 
As a control, we also reported curves obtained by inserting or deleting randomly selected features (gray). 
The shaded regions indicate variability across samples, corresponding to ±3 standard deviations.
Performance was quantified using the area under the curve (AUC).
Since all benchmark tasks considered in this study are binary classification tasks, the prediction probability was evaluated within the range [0.5,1.0], where 0.5 corresponds to random performance.
The results showed that the insertion curves consistently achieved higher AUC than the random baseline, while the deletion curves yielded lower AUC, demonstrating a sharper rise (or drop) in prediction confidence for small $k$. 
Furthermore, the differences between the proposed method and the random baseline were statistically significant under the Wilcoxon signed-rank test ($p<0.001$).
These findings provide quantitative evidence that the features highlighted by the proposed explanations are not merely correlational but are causally relevant to the model’s predictions.

\begin{table}[t]
    \centering
    \caption{
        Ablation study on frequency-domain feature extraction methods in the Wavelet-Convolution module. 
        We compared variants using STFT or CWT in the E-Branch and 1D convolutions in the A-Branch.
        Replacing the full configuration (1-v) with CWT-based (1-ii) and convolution-based (1-iii) variants reduced accuracy indicating the advantage of combining deterministic and adaptive frequency features.
        1D-Conv. denotes 1D convolutional layers.
    }
    \label{tab:ablation-wavelet}
    \renewcommand{\arraystretch}{1}
    \resizebox{\textwidth}{!}{
    \begin{tabular}{c c c c c c c}
      \toprule
      \multirow{2}{*}{\bf{Model}} & \multirow{2}{*}{\bf{E-Branch}} & \multirow{2}{*}{\bf{A-Branch}} & \multicolumn{2}{c}{\bf{OpenBMI~\cite{OpenBMI}}} & \multicolumn{2}{c}{\bf{Stieger2021~\cite{Stieger}}}\\
      \cmidrule(l{1mm}r{1mm}){4-5}
      \cmidrule(l{1mm}r{1mm}){6-7}
      & & & \bf{Accuracy[\%]$~\uparrow$} & \bf{Macro-F1[\%]$~\uparrow$} & \bf{Accuracy[\%]$~\uparrow$} & \bf{Macro-F1[\%]$~\uparrow$}\\
      \midrule
      (1-i)  & STFT & ---        & 71.51\std\small3.94  & 71.31\std\small3.95 & 80.05\std\small4.15 & 80.03\std\small4.17\\
      (1-ii) & CWT  & ---        & 78.86\std\small5.05  & 78.82\std\small5.07 & 83.66\std\small4.72  & 83.56\std\small4.81\\
      (1-iii)& ---  & 1D-Conv.   & 80.25\std\small5.55  & 80.15\std\small5.64  & 86.52\std\small4.51  & 86.51\std\small4.51\\
      (1-iv) & STFT & 1D-Conv.   & 79.46\std\small5.22  & 79.34\std\small5.31 & 85.87\std\small4.78 & 85.84\std\small4.82\\
      (1-v)  & CWT  & 1D-Conv.   & \bf{81.62}\std\bf{\small5.20} & \bf{81.56}\std\bf{\small5.24} & \bf{87.12}\std\bf{\small4.33} & \bf{87.11}\std\bf{\small4.33}\\
      \bottomrule
    \end{tabular}
    }
    \vspace{-1em}
\end{table}
\begin{table}[t]
    \centering
    \caption{
    Ablation study on architectures for capturing temporal dependencies.
    We compared (2-i) Attention~\cite{Transformer}, (2-ii) S4-LegS~\cite{S4}, (2-iii) Mega~\cite{Mega}, (2-iv) Mamba-2~\cite{Mamba2}, and (2-v) S5~\cite{S5} (as the proposed method) for modeling temporal dependencies.
    All alternatives underperformed Model (2-v) indicating the effectiveness of S5.
    }
    \label{tab:ablation-ssm}
    \renewcommand{\arraystretch}{1}
    \resizebox{\textwidth}{!}{
    \begin{tabular}{c c c c c c}
      \toprule
      \multirow{2}{*}{\bf{Model}} & \multirow{2}{*}{\bf{Architecture}} &\multicolumn{2}{c}{\bf{OpenBMI}~\cite{OpenBMI}} &\multicolumn{2}{c}{\bf{Stieger2021}~\cite{Stieger}}\\
      \cmidrule(l{1mm}r{1mm}){3-4}
      \cmidrule(l{1mm}r{1mm}){5-6}
      & &\bf{Accuracy [\%]$~\uparrow$} &\bf{Macro-F1[\%]$~\uparrow$} &\bf{Accuracy [\%]$~\uparrow$} &\bf{Macro-F1[\%]$~\uparrow$}\\
      \midrule 
      (2-i)   & Attention~\cite{Transformer} &78.74\std\small5.11 &78.63\small{$\pm$ 5.18} &85.10\small{$\pm$ 4.36} &85.09\small{$\pm$ 4.35}\\
      (2-ii)  & S4-LegS~\cite{S4}  &79.69\std\small4.98 &79.61\std\small5.02 &86.19\std\small5.07 &86.17\std\small5.07\\      
      (2-iii) & Mega~\cite{Mega} &79.86\std\small2.05  &79.81\std\small12.08 &86.31\std\small5.26 &86.28\std\small5.28\\
      (2-iv)   & Mamba-2~\cite{Mamba2} &80.29\std\small3.69 &80.26\small{$\pm$ 3.71} &86.89\small{$\pm$ 4.60} &86.86\small{$\pm$ 4.61}\\
      (2-v)  & S5~\cite{S5}  &\bf{81.62}\std\bf{\small5.20} &\bf{81.56}\bf{\small{$\pm$ 5.24}} &\bf{87.12}\bf{\small{$\pm$ 4.33}} &\bf{87.11}\bf{\small{$\pm$ 4.33}}\\
      \bottomrule
    \end{tabular}
    }
    \vspace{-1em}
\end{table}

\begin{table}[t]
    \centering
    \caption{
    Ablation study of core modules.
    Four configurations were evaluated by excluding each component: (3-i) w/o Wavelet-Convolution, (3-ii) w/o Frequency-SSM, (3-iii) w/o Channel-SSM, and (3-iv) full model.
    Compared to Model (3-iv), all ablated variants show reduced accuracy, confirming the contribution of each component.
    }
    \label{tab:ablation-module}
    \renewcommand{\arraystretch}{1}
    \resizebox{\textwidth}{!}{
    \begin{tabular}{c c c c c c c c}
      \toprule
      \multirow{2}{*}{\bf{Model}} & \bf{Wavelet} & \bf{Frequency} & \bf{Channel} &\multicolumn{2}{c}{\bf{OpenBMI}~\cite{OpenBMI}} &\multicolumn{2}{c}{\bf{Stieger2021}~\cite{Stieger}}\\
      \cmidrule(l{1mm}r{1mm}){5-6}
      \cmidrule(l{1mm}r{1mm}){7-8}
      & \bf{Conv.} & \bf{SSM} & \bf{SSM} &\bf{Accuracy~[\%]$~\uparrow$} &\bf{Macro-F1~[\%]$~\uparrow$} &\bf{Accuracy~[\%]$~\uparrow$} &\bf{Macro-F1~[\%]$~\uparrow$}\\
      \midrule 
      (3-i)   &            & \checkmark & \checkmark &75.88\std\small2.12 &75.80\small{$\pm$2.13} &84.38\std\small3.49 &84.36\small{$\pm$3.50}\\
      (3-ii)  & \checkmark &            & \checkmark &79.14\std\small2.75 &79.11\small{$\pm$ 2.77} &84.52\small{$\pm$ 3.21} &84.50\small{$\pm$ 3.22}\\
      (3-iii) & \checkmark & \checkmark &            &80.64\std\small2.31 &80.61\small{$\pm$ 2.32} &85.74\small{$\pm$ 3.14} &85.73\small{$\pm$ 3.14}\\
      (3-iv)  & \checkmark & \checkmark & \checkmark &\bf{81.62}\std\bf{\small5.20} &\bf{81.56}\bf{\small{$\pm$ 5.24}} &\bf{87.12}\std\bf{\small4.33} &\bf{87.11}\bf{\small{$\pm$ 4.33}}\\
      \bottomrule
    \end{tabular}
    }
    \vspace{-1em}
\end{table}

\subsection{Ablation Study}
\label{section:ablation}
To investigate the effectiveness of each module, we conducted ablation studies on the following three conditions.

\paragraph{Wavelet-Convolution ablation.}
Table \ref{tab:ablation-wavelet} presents the performance of different frequency domain feature extraction methods in the E-Branch and A-Branch of the Wavelet-Convolution module. 
We compared five model variants (1-i)-(1-v) with combinations of short-time Fourier transform (STFT) and continuous wavelet transform (CWT) in the E-Branch, along with 1D convolutional layers in the A-Branch. 
The results reveal that models employing CWT in the E-Branch exhibited markedly different performance on OpenBMI~\citep{OpenBMI}, with the classification accuracy of Model~(1-ii) being 2.76 points lower than that of Model~(1-v).
Similarly, when models using 1D convolutions in the A-Branch were compared, Model~(1-iii) underperformed Model~(1-v) by 1.37 points in the corresponding metric.

To explain this observation, we consider the intrinsic properties of MI EEG signals, which exhibit both well-defined frequency characteristics and pronounced non-stationarity.
Regarding frequency characteristics, MI EEG signals contain discriminative patterns in specific frequency bands, particularly the $\mu$-band (8–12 Hz), as widely reported in prior work~\cite{pfurtscheller06} and confirmed by our visual explanations (Section 6.2).
Accordingly, employing the Continuous Wavelet Transform (CWT) in the E-Branch enables deterministic extraction of such physiologically meaningful components.
Moreover, the superiority of the CWT-based configuration (Model (1-v)) over the STFT-based variant (Model (1-iv)) can be attributed to the transient nature of $\mu$-band modulations, which are not well captured by the fixed temporal resolution of STFT.
Furthermore, in terms of non-stationarity, EEG signals are affected by temporal electrode shifts, device characteristics, and subject-specific variability~\cite{xiong2025}.
To account for these factors, the A-Branch employs one-dimensional convolutional layers that function as adaptive filters, enabling the model to learn task-relevant frequency patterns while compensating for such variations.
Taken together, these results suggest that the integration of deterministic frequency decomposition (via CWT) and adaptive feature learning (via one-dimensional convolution) provides a complementary inductive bias that is well aligned with the characteristics of MI EEG signals.

\paragraph{Frequency-SSM and Channel-SSM ablation.}
Table~\ref{tab:ablation-ssm} shows the performance of different architectures in the Frequency-SSM and Channel-SSM. 
We compared models using the following architectures for capturing temporal dependencies: (2-i) Attention~\citep{Transformer}, (2-ii) S4-LegS~\citep{S4}, (2-iii) Mega~\citep{Mega}, (2-iv) Mamba-2~\citep{Mamba2}, and (2-v) S5~\citep{S5}. 
Table~\ref{tab:ablation-ssm} indicates that the classification accuracy of models (2-i), (2-ii), (2-iii), and (2-iv) on OpenBMI underperformed Model~(2-v) by 2.88, 1.93, 1.76, and 1.33 points, respectively.
These results suggest that S5, a Deep SSM employing a time-invariant system and MIMO configuration, is effective for capturing temporal dependencies in EEG signals.

\paragraph{Module-wise ablation.}
Table~\ref{tab:ablation-module} presents the performance of the three main modules: Wavelet-Convolution, Frequency-SSM, and Channel-SSM. 
We compared four model configurations: (3-i) exclusion of Wavelet-Convolution, (3-ii) exclusion of Frequency-SSM, (3-iii) exclusion of Channel-SSM, and (3-iv) the complete model incorporating all three modules. 
Regarding classification accuracy on OpenBMI, Models~(3-i), (3-ii), and (3-iii) underperformed Model~(3-iv) by 5.74, 2.48, and 0.98 points, respectively.
These findings indicate that each module contributes to improving overall model performance, with the Wavelet-Convolution module exerting the most significant impact.

\subsection{Sensitivity Analysis}

\paragraph{Sequence length analysis}
\begin{figure}[t]
    \centering
    \includegraphics[width=0.9\linewidth]{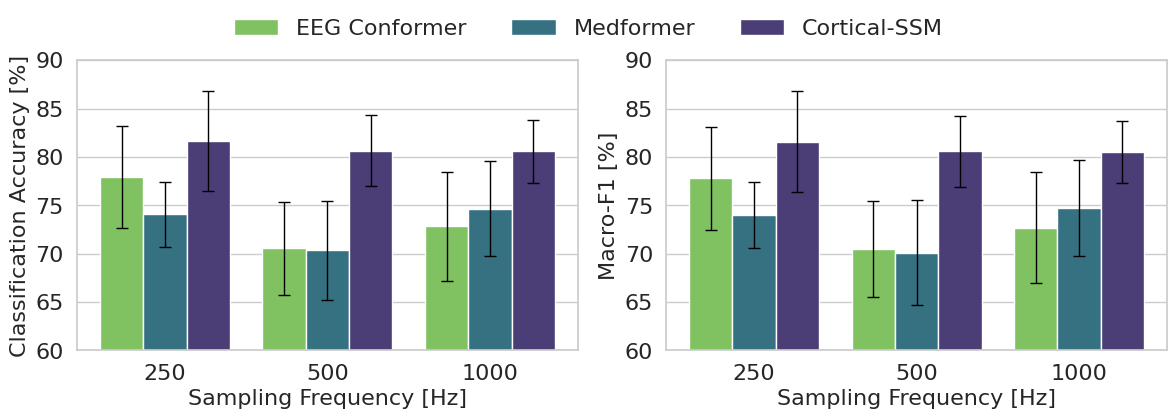}
    \vspace{-1em}
    \caption{
    Sensitivity to sequence length on OpenBMI~\citep{OpenBMI}.
    EEG signals (4~s segments) were resampled to 250, 500, and 1000~Hz.
    While baselines revealed notable classification accuracy degradation (up to -7.34 and -3.73 points), the proposed method exhibited only a marginal drop (-1.06 point), indicating improved scalability with respect to sequence length.
    }
    \label{fig:seq_len}
\end{figure}
To investigate the effect of sequence length on decoding performance, we performed a sensitivity analysis on the OpenBMI dataset by resampling EEG signals originally recorded at 1000 Hz to three sampling rates (250 Hz, 500 Hz, and 1000 Hz), as illustrated in Figure~\ref{fig:seq_len}.
The 250 Hz condition corresponds to the setting used in the original experimental setup, and all conditions employ 4-second EEG segments as input.
We compared our method against two representative EEG decoding baselines: EEG Conformer~\citep{EEGConformer} and Medformer~\citep{Medformer}.
While EEG Conformer and Medformer revealed accuracy drops of up to 7.34 and 3.73 points, respectively, our method shows only a marginal decrease of 1.06 points.
These results suggest that the proposed method scales with sequence length.

\paragraph{Signal-to-noise ratio ablation}
\begin{figure}[t]
    \centering
    \includegraphics[width=0.9\linewidth]{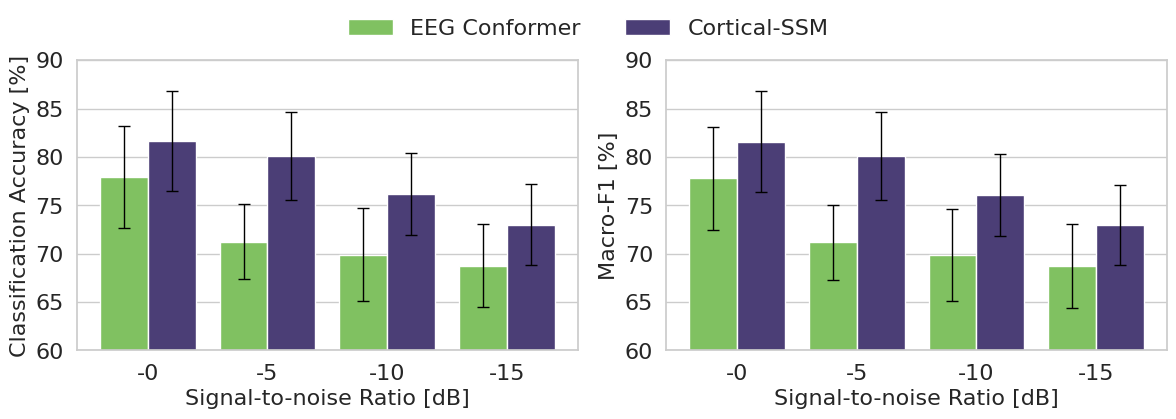}
    \vspace{-1em}
    \caption{
    SNR was progressively degraded (-0, -5, -10, -15~dB) by injecting noise into band-limited (1-100~Hz) EEG signals.
    The proposed method consistently outperformed baselines across all SNR levels in terms of classification accuracy and macro-F1.
    }
    \label{fig:snr}
\end{figure}
To evaluate the impact of noise and signal degradation, we conducted an additional experiment in which the signal-to-noise ratio (SNR) of EEG signals from the OpenBMI dataset was progressively degraded by -0~dB, -5~dB, -10~dB, and -15~dB as shown in Figure~\ref{fig:snr}.
Here, we compared the performance of our proposed method with the second-best baseline, EEG Conformer~\citep{EEGConformer}. 
The signal bandwidth was set to 1-100~Hz. 
As shown in the figure, our method consistently outperformed EEG Conformer across all SNR levels in terms of classification accuracy and macro-F1, demonstrating its effectiveness even in the presence of substantial noise and signal degradation common in clinical settings.

\section{Discussion}
While the results presented in Table~\ref{tab:quantitative_main} indicate the potential advantages of the proposed method, it is important to note that the current formulation assumes a synchronous setting.
In this study, synchronous data with explicitly controlled motor imagery timing were adopted to enable principled interpretation of the visual explanations. 
Because the onset of motor imagery is known in this setting, the temporal correspondence between model attributions and the underlying neural processes can be clearly interpreted.
In practical applications, however, MI EEG decoding is typically required to operate in an asynchronous setting, where the onset of motor imagery is unknown.
Extending the proposed method to support interpretable visual explanations under such asynchronous conditions therefore remains an important direction for future work.

In addition to this experimental constraint, as with the existing baselines, Cortical-SSM does not explicitly address subject-level domain shifts~\citep{Cho,Kaya}. 
As shown in Section~\ref{section:quantitative}, the Shapiro–Wilk test rejects normality across folds, suggesting substantial variance.
Incorporating domain adaptation techniques, such as learning invariant representations offers a promising path forward.

From the architectural perspective, Cortical-SSM primarily fuses multi-domain features at a later stage of the model, following largely independent processing in earlier stages.
While this design enhances interpretability, it may in some cases encourage greater reliance on a subset of domains (e.g., temporal) when they are sufficient for the task, potentially limiting the utilization of complementary information from other domains. 
Future work could explore more tightly coupled or progressive integration strategies to further promote balanced feature utilization. 

Moreover, although the visual explanations of the proposed method highlight brain regions that are broadly consistent with established neurophysiological findings, caution is required when interpreting these patterns in causal terms. 
The visualization reflects gradients of the prediction error and does not directly measure neural activation itself. 
Accordingly, the generated explanations should be understood as emphasizing discriminative patterns that contribute to accurate MI classification, rather than as providing direct evidence of underlying neurophysiological mechanisms.
Given this limitation, future work may also explore integrating the proposed method with Electrophysiological Source Imaging (ESI) approaches~\cite{sLORETA, DeepSIF}, enabling a more direct linkage to cortical sources and providing insights into the causal structure of the decoding process.

Despite these considerations, the present findings suggest that jointly modeling the spatial, temporal, and spectral structure of EEG is a useful strategy for MI EEG decoding. 
Moreover, providing visual explanations appears particularly important for EEG-based brain functional imaging, where the inherently low signal-to-noise ratio makes physiologically plausible interpretation essential for reliable decoding.

\section{Conclusion}
\label{section:conclusion}
In this study, we focused on a classification task based on EEG signals recorded during MI tasks.
We proposed Cortical-SSM, an extension of Deep SSMs designed to capture integrated dependencies across temporal, spatial, and frequency domains.
For frequency feature extraction, we proposed the Wavelet-Convolution, which extracts non-black-box frequency-analyzable features while maintaining a learnable representation.
In comprehensive evaluations on two MI EEG benchmarks, our method consistently outperformed the comparison baseline methods. 
Furthermore, we demonstrated that for both EEG signals, neurophysiologically significant regions were attended to in the visual explanations generated by our proposed method.

\data{
The data that support the findings of this study are publicly available.
Specifically, the datasets used in this study can be accessed from the following sources:
\begin{itemize}
    \item OpenBMI~\cite{OpenBMI}: \url{https://gigadb.org/dataset/100542}
    \item Stieger2021~\cite{Stieger}: \url{https://figshare.com/articles/dataset/Human_EEG_Dataset_for_Brain-Computer_Interface_and_Meditation/13123148}
\end{itemize}
No new data were generated in this study.
}

%


\funding{
This work was partially supported by JSPS KAKENHI Grant Number 23K28168 and JST Moonshot.
}




\bibliographystyle{unsrt}
\bibliography{refs}

\newpage
\end{document}